\documentclass[conference]{IEEEtran}
\IEEEoverridecommandlockouts
\usepackage{cite}
\usepackage{amsmath,amssymb,amsfonts}
\usepackage{algorithmic}
\usepackage{graphicx}
\usepackage{textcomp}
\usepackage{xcolor}
\usepackage{etoolbox}
\usepackage{float}
\usepackage{xcolor}
\usepackage[hyphens]{url}
\usepackage{hyperref}
\usepackage{amssymb}
\usepackage{url}
\usepackage{multirow}
\usepackage{color, colortbl}
\usepackage{subcaption, booktabs}
\usepackage{csquotes}
\usepackage{xspace}
\usepackage{wrapfig}
\usepackage{fontawesome}
\usepackage[leftcaption]{sidecap}

\definecolor{tabhighlight}{HTML}{e5e5e5}
\def\ie{\emph{i.e.}\xspace}

\newcommand{\rotbox}[1]{\rotatebox{55}{#1}}

\def\BibTeX{{\rm B\kern-.05em{\sc i\kern-.025em b}\kern-.08em
    T\kern-.1667em\lower.7ex\hbox{E}\kern-.125emX}}
\begin{document}

\title{MuDPT: Multi-modal Deep-symphysis Prompt Tuning for Large Pre-trained Vision-Language Models}

\author{\IEEEauthorblockN{Yongzhu Miao, Shasha Li$^{\ast}$, Jintao Tang$^{\ast}$ and Ting Wang$^{\ast}$ \thanks{*Corresponding author.}}
\IEEEauthorblockA{\textit{College of Computer Science and Technology} \\
\textit{National University of Defense Technology}\\
Changsha, P.R.China \\
\{miaoyz, shashali, tangjintao, tingwang\}@nudt.edu.cn}}

\maketitle

\begin{abstract}
Prompt tuning, like CoOp, has recently shown promising vision recognizing and transfer learning ability on various downstream tasks with the emergence of large pre-trained vision-language models like CLIP. However, we identify that existing uni-modal prompt tuning approaches may result in sub-optimal performance since this uni-modal design breaks the original alignment of textual and visual representations in the pre-trained model. Inspired by the nature of pre-trained vision-language models, we aim to achieve completeness in prompt tuning and propose a novel approach called \textbf{Mu}lti-modal \textbf{D}eep-symphysis \textbf{P}rompt \textbf{T}uning, dubbed as \textbf{MuDPT}, which extends independent multi-modal prompt tuning by additionally learning a model-agnostic transformative network to allow deep hierarchical bi-directional prompt fusion. We evaluate the effectiveness of MuDPT on few-shot vision recognition and out-of-domain generalization tasks. Compared with the state-of-the-art methods, MuDPT achieves better recognition and generalization ability with an apparent margin thanks to synergistic alignments of textual and visual representations. 
\end{abstract}


\section{Introduction}
Conventional visual representation learning is training vision models to predict a fix-set of visual objects with pre-determined categories \cite{b1}. Nevertheless, this restricted paradigm limits the representation to closed-set visual concepts and makes resulting models inferior generalization ability to the objects of never-seen categories \cite{b2}. More recently, large-scale vision-language models pre-trained on massive amounts of image-text pairs, known as VL-PTMs, have been a promising alternative and shown conspicuous zero-shot transferability to diverse downstream tasks, such as SimVLM \cite{b3}, CLIP \cite{b4}, ALIGN \cite{b5}. These VL-PTMs are capable of representing open-set visual concepts thanks to a broader source of supervision contained in natural language contexts.

\begin{figure}[ht]
    \centering
    \includegraphics[width=0.94\linewidth]{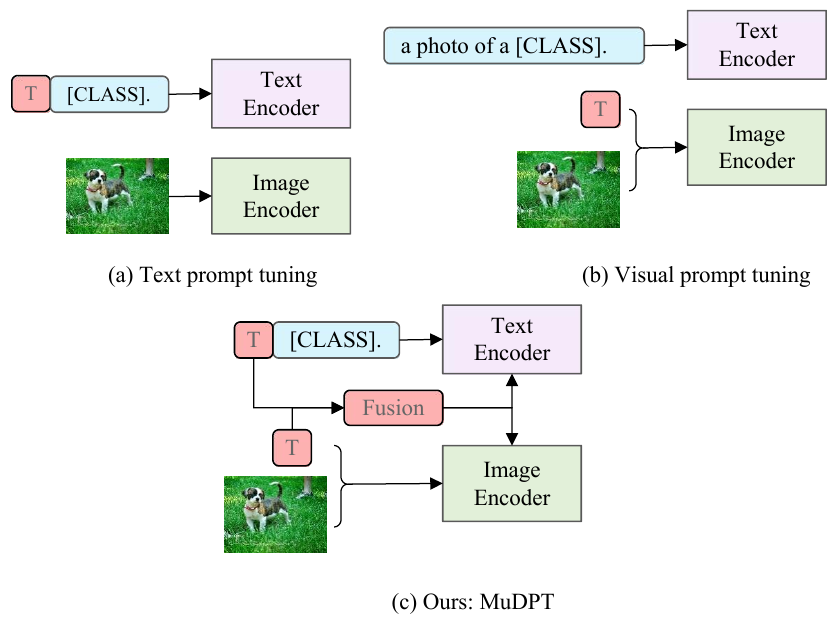}
    \caption{Comparison between our approach (MuDPT) and existing prompt tuning approaches (text prompt tuning and visual prompt tuning).}
    \label{fig:compare}
\end{figure}

With the VL-PTMs demonstrating attractive generalization capability, the community sets out to explore optional solutions for migrating these models to downstream tasks \cite{b6, b7, b8, b9}, e.g., visual grounding \cite{b6}, visual entailment \cite{b7}, visual recognition \cite{b8}. An effective approach is to mitigate the gap between the form of pre-training and downstream task by leveraging carefully determined manual prompts (\emph{e.g.}, “a photo of a [CLASS].”) injected into pre-trained models \cite{b4}. Nevertheless, manual prompt engineering is an intuitive and experience-oriented work, and usually issues in sub-optimal performance \cite{b8}. Some prompt tuning models are developed for automated prompt engineering, such as CoOp \cite{b8}, CoCoOp \cite{b10}, VPT \cite{b11}, etc. However, as shown in Fig. \ref{fig:compare}, the previous work has focused on uni-modal prompt tuning in textual or visual branch of VL-PTMs, while multi-modal prompt tuning in the two modalities simultaneously remains an unexplored area in the literature. 

In this paper, we identify that none of the uni-modal prompt tuning methods leads to consistent performance improvements. We assume that the potential reason could be the corruption of already aligned textual and visual representations in the VL-PTMs in uni-modal prompt tuning. This problem can also be observed in Fig. \ref{fig:compare}. In existing uni-modal prompt tuning methods, one of the modalities of a VL-PTM is completely frozen during downstream training. For instance, in text prompt tuning (Fig. \ref{fig:compare} (a)), the textual representation is dynamically optimized by introducing learnable parameters to the text encoder, while the opposite visual representation is completely fixed. This uni-modal design defeats the nature of VL-PTMs, where the intention is to mutually align textual and visual representations in a common multi-modal space \cite{b4}. Previous researches have proven that alignment between multi-modal representations can enhance the performance of VL-PTMs on downstream tasks \cite{b5}. 

Starting from the intrinsic intent of CLIP pre-training, where synergistically training text encoder and image encoder to align textual and visual representations \cite{b4}, our motivation is to introduce both textual and visual deep prompts for dynamically coordinating textual and visual representations. An intuitionistic solution is feeding multi-modal prompts into encoders respectively and performing end-to-end supervised training to optimize the prompts individually. However, our early attempts verify that such a naive joint prompt tuning generally leads to sub-optimal performance due to the essential discrepancy between the two modalities. 

To achieve completeness in prompt tuning approaches for better adaption of VL-PTMs, we present \textbf{Mu}lti-modal \textbf{D}eep-symphysis \textbf{P}rompt \textbf{T}uning, \textbf{MuDPT}, to adequately fine-tune the textual and visual representations by further learning a modality-agnostic transformative network. We evaluate the effectiveness of our approach on few-shot visual recognition and out-of-domain generalization tasks. Compared with current state-of-the-art methods: CoOp \cite{b8} and CoCoOp \cite{b10}, we obtain better few-shot visual recognition performance with an obvious margin. On base-to-new generalization experiment, our method \textbf{MuDPT} achieves an absolute average gain of 11.28\% on new classes and 6.7\% on harmonic-mean compared CoOp. Further, MuDPT demonstrates beneficial transferability in cross-dataset evaluation and domain generalization settings, resulting in better trade-off compared with existing approaches. Extensive experiments demonstrates that MuDPT achieves better few-shot visual recognition and generalization performance thanks to the synergistic alignment of textual and visual representations during prompt tuning on downstream datasets. 

Our main contributions are summarized below:
\begin{itemize}
\item We propose \textbf{Mu}lti-modal \textbf{D}eep-symphysis \textbf{P}rompt \textbf{T}uning (\textbf{MuDPT}) for CLIP to realize synergistic textual and visual representation alignment dynamically, which is the first multi-modal deep synergistic prompt tuning to the best of our knowledge. 
\item To achieve cross-modality prompt transformation and fusion, We propose a light modality-agnostic transformative network. It builds the synergistic bridge between text and image prompts.
\item We conduct extensive experiments to validate the performance of our approach and show that \textbf{MuDPT} is a practical strategy that outperforms the existing approaches. 
\end{itemize}

\section{Methodology}
\begin{figure*}[ht]
\centering
\includegraphics[width=0.9\linewidth]{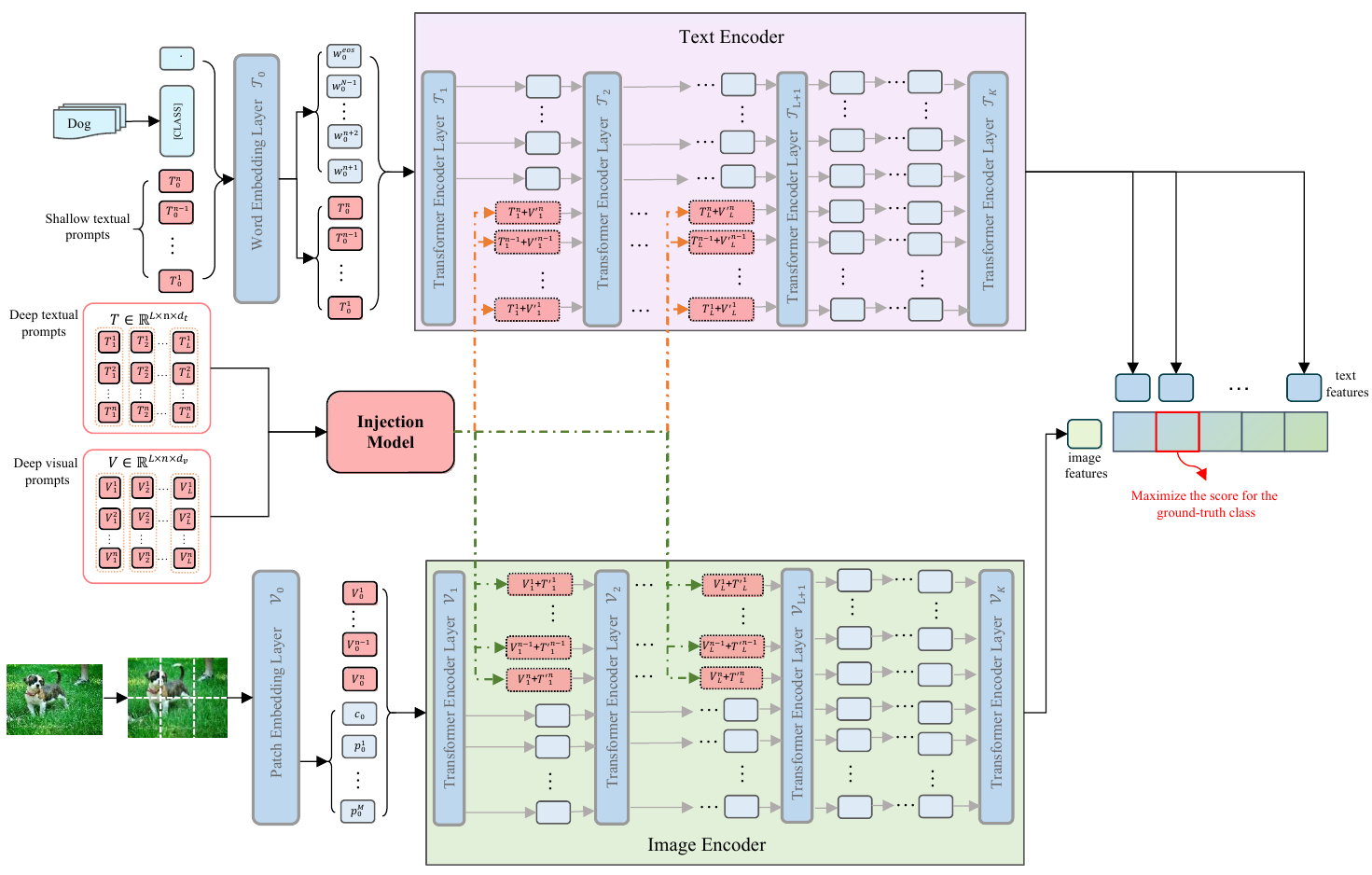}
\caption{Overview of our approach: \textbf{MuDPT} (\textbf{Mu}lti-modal \textbf{D}eep-symphysis \textbf{P}rompt \textbf{T}uning). MuDPT introduces textual and visual prompts to the text and image encoder. During training, only the parameters of prompts (drawn in pink blocks) are tuned while the backbone (drawn in blue blocks) is frozen. MuDPT realizes cross-modality prompt transformation and fusion by further learning a light \textbf{Injection Model}. }
\label{fig:mudpt}
\end{figure*}

Our approach concerns multi-modal deep-symphysis prompt tuning to adapt VL-PTMs, \emph{e.g.}, CLIP, for better visual recognition and generalization performance to downstream tasks. Fig. \ref{fig:mudpt} shows an overview of our approach. Firstly, We symmetrically introduce deep learnable textual and visual prompts $T$ and $V$ for the textual and visual branches of CLIP separately. Further, we design a transformative block, called \textbf{Injection Model}, to establish the cross-modality attention between the textual and visual prompts for deep hierarchical bi-directional prompt fusion. In this section, we first briefly introduce the architecture of CLIP in Sec. \ref{sec:preliminaries}. Then, we present the technical rationale of our proposed MuDPT in Sec. \ref{sec:mudpt}.

\subsection{Preliminaries}
\label{sec:preliminaries}

\begin{figure}[ht]
\centering
\includegraphics[width=0.95\linewidth]{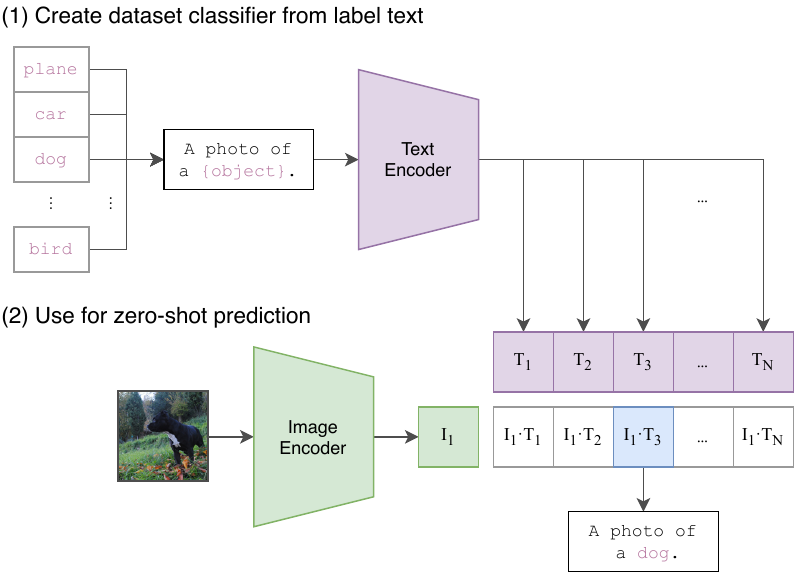}
\caption{Overview of CLIP. }
\label{fig:clip}
\end{figure}

As shown in Fig. \ref{fig:clip}, \textbf{CLIP}\cite{b4} is a Transformer-based VL-PTM proposed by Radford et al., which consists of an image encoder and a text encoder. Please refer to \cite{b4} for more details. We utilize a ViT-based version of CLIP, which is also consistent with existing prompt tuning approaches \cite{b8, b10}. 

\textbf{Text Encoder} $\mathcal{T}$ of CLIP takes a sequence of word embeddings $W_0 = \{w^1_0, w^2_0, ..., w^{N-1}_0, w^{eos}_0\} \in \mathbb{R} ^ {N \times d_t}$ as input, where $w^i_0$ denotes the word embedding of a single token with the dimension of $d_t$. Further, the text encoder feeds the word embeddings $W_0$ into a Transformer $\mathcal{T}$. At each hidden layer of the Transformer, the output of the previous layer $W_i$ sequentially serves as the input to the next layer $\mathcal{T}_{i+1}$: 
\begin{eqnarray}
W_{i+1}=\mathcal{T}_{i+1}(W_i), {i \in [0, K)}
\end{eqnarray}
Here, the Transformer of the text encoder has $K$ hidden layers. The final vectorized representation $z$ of input text is obtained by mapping the last hidden state corresponding to the $[EOS]$ token $w^{eos}_K$ in the last Transformer block to a common multi-modal space 
via $TextProj \in \mathbb{R} ^ {d_t \times d_c}$: 
\begin{eqnarray}
\label{eq:2}
z=w_K^{eos} \circ TextProj
\end{eqnarray}
Where $eos$ is the index of $[EOS]$ token in the input sequence and $(\cdot \circ \cdot)$ denotes matrix multiplication, ${d_c}$ is the dimension of the common multi-modal space.

\textbf{Image encoder} $\mathcal{V}$ firstly encodes the image $I$ into fixed-sized patch embeddings 
$P_0 = \{p^1_0, p^2_0, ..., p^M_0\} \in \mathbb{R}^{M \times d_v}$. The patch embeddings with an attached learnable $[CLS]$ token embedding $c_0 \in \mathbb{R}^{d_v}$ then input to a visual Transformer $\mathcal{V}$ to obtain the last hidden states:
\begin{eqnarray}
\label{eq:3}
[c_{i+1}, P_{i+1}]=\mathcal{V}_{i+1}([c_i, P_i]), {i \in [0, K)}
\end{eqnarray}
Here, $[\cdot, \cdot]$ denotes concatenation operation. To get the feature vector $x$ of input image $I$, the last class embedding of the $[CLS]$ token $c_K$ in the last Transformer block is projected to a common multi-modal space via $ImageProj \in \mathbb{R} ^ {d_v \times d_c}$: 
\begin{eqnarray}
\label{eq:4}
x=c_K \circ ImageProj
\end{eqnarray}

\textbf{Inference}. For visual recognition, a manual text prompt with class symbol, \emph{e.g.}, "a photo of a [CLASS]", is injected into the text encoder to synthesize classification weights $Z = \{z_1, z_2, ..., z_m\}$ by filling in "[CLASS]" with the specific class name. The prediction $\hat{y}$ of a given image $I$ having the highest cosine similarity between the classification weight $z_{\hat{y}}$ and image representation $x$, as the following form: 
\begin{eqnarray}
\label{eq:5}
p(\hat{y}|x)= \frac{exp(<x, z_{\hat{y}}>/\tau)}{\sum_{i=1}^{m} exp(<x, z_i>/\tau)}
\end{eqnarray}
where $m$ is the number of categories and $\tau$ is a fixed temperature value which is optimized during pre-training. 

\subsection{Multi-modal Deep-symphysis Prompt Tuning}
\label{sec:mudpt}

\begin{figure*}[ht]
\centering
\includegraphics[width=0.6\linewidth]{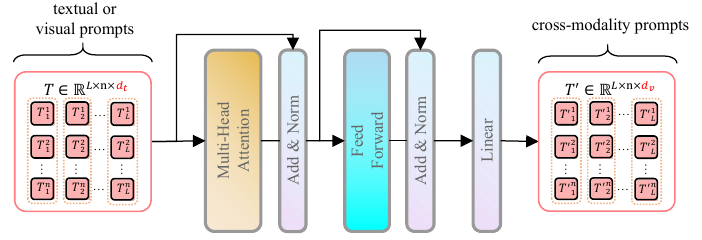}
\caption{Overview of our Injection Model, which consists of a multi-head attention block to calculate cross-modality attention and a linear layer to adapt the dimension of prompts.} 
\label{fig:inj}
\end{figure*}

Driven by the nature of CLIP, we propose \textbf{Mu}lti-modal \textbf{D}eep-symphysis \textbf{P}rompt \textbf{T}uning, \textbf{MuDPT}, to achieve completeness in prompt tuning for better adapting VL-PTMs to visual recognition task. We follow the workflow of CLIP \cite{b4}. 

As shown in Fig. \ref{fig:mudpt}, for deep hierarchical bi-directional prompt fusion, we design a light transformative block, \textbf{Injection Model}, to establish the cross-modality attention between multi-modal prompts, which is described in Sec. \ref{sec:dspt}. In addition to only introduce learnable prompts in the embedding layer like CoOp \cite{b8}, we further inject other groups of prompts into deeper Transformer layers to progressively model stage-wise textual and visual representations, which are narrated in Sec. \ref{sec:dtpt} and Sec. \ref{sec:dvpt} separately. 

\subsubsection{Deep-symphysis Prompt Fusion}
\label{sec:dspt}
From the consideration of establishing the cross-modality interaction, we propose a tiny transformative block, called \textbf{Injection Model}, to share prompts across two modalities, as shown in the middle of Fig. \ref{fig:mudpt}. The hidden layers of Injection Model are illustrated in Fig. \ref{fig:inj}. 

Deep textual prompts $T\in \mathbb{R}^{L \times n \times d_t}$ and visual prompts $V \in \mathbb{R}^{L \times n \times d_v}$ are introduced in the textual and visual branches of CLIP separately, as illustrated in Fig. \ref{fig:mudpt}. Here, $L$ is a hyper-parameter indicating the depth of prompt and $n$ is the length of prompts. To ensure hierarchical bi-directional prompt fusion, we further take textual prompts $T$ and visual prompts $V$ as inputs into the \textbf{Injection Model} and obtain cross-modality prompts $T'$ and $V'$: 
\begin{eqnarray}
T', V' &=& Injection(T, V)
\end{eqnarray}
where $Injection$ means the hidden layers in the Injection Model, which are shown in Fig. \ref{fig:inj}. Then multi-modal prompts $\hat{T}$ and $\hat{V}$ are now composed of original prompts and cross-modality prompts, which we called hierarchical bi-directional prompt fusion, as formulated in the following:
\begin{eqnarray}
\hat{T}, \hat{V} &=& T + V', V + T'
\end{eqnarray}
Based on this modification, we demonstrate how to perform multi-modal deep-symphysis prompt tuning in Sec. \ref{sec:dtpt} and Sec. \ref{sec:dvpt}. 

\subsubsection{Deep Text Prompt Tuning}
\label{sec:dtpt}
We conduct deep text prompt tuning in the language branch of CLIP, seen in the upper part of Fig. \ref{fig:mudpt}. In our approach, the input embedding is the concatenation of textual prompts in the embedding layer $T_0 \in \mathbb{R}^{n \times d_t}$ and the word embedding of a specific class name, which is reformulated as $[T_0, W_0, w^{eos}_0] \in\mathbb{R}^{N \times d_t}$ where ${W_0=[w^{n+1}_0, ..., w^{N-1}_0]}$ is the word embedding of the class name and $w^{eos}_0$ is the embedding of $[EOS]$ token. Textual prompts and input embedding are then 
further injected into each Transformer layer up to $L^{th}$ layer: 
\begin{eqnarray}
\label{eq:8}
[\_, W_{i+1}, w^{eos}_{i+1}] = 
\left\{
  \begin{aligned}
  &\mathcal{T}_{i+1}([T_{i}, W_{i}, w^{eos}_{i}]), i = 0 \\
  &\mathcal{T}_{i+1}([\hat{T}_{i}, W_{i}, w^{eos}_{i}]), i \in [1, L)
  \end{aligned}
\right.
\end{eqnarray}
After the $L^{th}$ layer, the subsequent layers sequentially process the output of the previous layer, and the final text representation $z$ is calculated by the last hidden state of $w^{eos}_{K}$ and $TextProj$ as (\ref{eq:2}):
\begin{eqnarray}
[\hat{T}_{j+1}, W_{j+1}, w^{eos}_{j+1}] &=& \mathcal{T}_{j+1}([\hat{T}_{j}, W_j, w^{eos}_{j}])
\end{eqnarray}
Here, $j \in [L, K)$. When $L=1$, the learnable vectors are only introduced in the embedding layer of the text branch of CLIP, and deep text prompt tuning degrades into CoOp \cite{b8}.

\subsubsection{Deep visual prompt tuning}
\label{sec:dvpt}
Oppositely, we introduce deep visual prompts in the visual branch of CLIP, as shown in the bottom of Fig. \ref{fig:mudpt}. When we apply multi-modal prompt parameters $\hat{V} \in \mathbb{R}^{L \times n \times d_v}$ to the image encoder of CLIP, the class embedding of $[CLS]$ token $c_K$ in the last Transfomer layer is now calculated as follows:
\begin{eqnarray}
[c_{i+1}, \_, P_{i+1}] =
\left\{
  \begin{aligned}
  &\mathcal{V}_{i+1}([c_i, V_i, P_i]), i = 0 \\
  &\mathcal{V}_{i+1}([c_i, \hat{V}_i, P_i]), i \in [1, L)
  \end{aligned}
\right.
\end{eqnarray}
After the $L^{th}$ layer, the subsequent layers take similar processes: 
\begin{eqnarray}
\label{eq:11}
[c_{j+1}, \hat{V}_{j+1}, P_{j+1}] & = & \mathcal{V}_{j+1}([c_j, \hat{V}_j, P_j])
\end{eqnarray}
where, $j \in [L, K)$. The final image representation $x$ is given by mapping the last hidden state of class embedding $c_K$ to the common multi-modal space via $ImageProj$ as (\ref{eq:4}). We utilize the cosine similarity between the two representations to obtain the prediction $\hat{y}$ as illustrated in (\ref{eq:5}). We perform cross-entropy loss between ground-truth $y$ and prediction $\hat{y}$ to optimize the parameters of multi-modal prompts and the Injection Model. The learning objective is to maximize the predictive score for the ground-truth class. 

\section{Experiments}

\begin{table*}[h!]
\centering
\caption{Comparison of MuDPT with zero-shot CLIP, CoOp, and CoCoOp on few-shot visual recognition. Overall, MuDPT achieves the highest average accuracy over 11 datasets, indicating MuDPT is a better few shot visual recognition learner. }
\label{tab:imcls}
\resizebox{0.7\linewidth}{!}{
\begin{tabular}{l c ccccccccccc}
\toprule
& \multicolumn{11}{c}{\textbf{Dataset}} \\ \cmidrule(lr){2-13}
\textbf{Method}& \rotbox{ImageNet} & \rotbox{Caltech101} & \rotbox{OxfordPets} & \rotbox{StanfordCars} & \rotbox{Flowers102} & \rotbox{Food101} & \rotbox{FGVCAircraft} & \rotbox{SUN397} & \rotbox{DTD} & \rotbox{EuroSAT} & \rotbox{UCF101} & \rotbox{\emph{Average}} \\
\midrule
CLIP    & 66.7 & 92.7 & 89.1 & 65.3 & 70.8 & 86.3 & 24.8 & 62.6 & 44.3 & 41.8 & 66.7 & 64.65 \\
CoOp    & 63.57 & 92.43 & 89.47 & 71.2 & 93.1 & 77.17 & 29.07 & 69.77 & 64.2 & 80.1 & 74.9 & 73.18 \\
CoCoOp  & 72.4 & 95.4 & 95.8 & 71.6 & 79.4 & 89.2 & 33 & 76.8 & 63 & 73.8 & 75.4 & 75.07 \\
\midrule
\rowcolor{tabhighlight} MuDPT & 72.7 & \textbf{96.72} & 94.19 & \textbf{80.43} & \textbf{96.47} & 87.38 & \textbf{43.13} & 76.18 & \textbf{72.37} & \textbf{92.17} & \textbf{83.53} & \textbf{81.38} \\
MuDPT vs. CoOp                & {+9.13} & {+4.29} & {+4.72} & {+9.23} & {\textbf{+3.37}} & {+10.21} & {\textbf{+14.06}} & {+6.41} & {+8.17} & {+12.07} & {+8.63} & {\textbf{+8.2}} \\ 
MuDPT vs. CoCoOp & ${\nearrow}$ & ${\nearrow}$ & ${\searrow}$ & ${\nearrow}$ & ${\nearrow}$ & ${\searrow}$ & ${\nearrow}$ & ${\searrow}$ & ${\nearrow}$ & ${\nearrow}$ & ${\nearrow}$ & {\textbf{+6.31}} \\ 
\bottomrule
\end{tabular}}
\end{table*}

\begin{table*}[h!]
  \centering
  \caption{Comparison of MuDPT with other prompt tuning methods on cross-dataset evaluation. Overall, MuDPT achieves the highest average accuracy over 10 datasets, indicating a better trade-off of performance between source and target datasets.}
  \label{tab:xd}
  \resizebox{0.7\linewidth}{!}{%
  \begin{tabular}{l c ccccccccccc}
  \toprule
  & \textbf{Source} & \multicolumn{11}{c}{\textbf{Target}} \\ \cmidrule(lr){2-2} \cmidrule(lr){3-13}
  & \rotbox{ImageNet} & \rotbox{Caltech101} & \rotbox{OxfordPets} & \rotbox{StanfordCars} & \rotbox{Flowers102} & \rotbox{Food101} & \rotbox{FGVCAircraft} & \rotbox{SUN397} & \rotbox{DTD} & \rotbox{EuroSAT} & \rotbox{UCF101} & \rotbox{\emph{Average}} \\
  \midrule
  CoOp    & 71.51 & 93.7 & 89.14 & 64.51 & 68.71 & 85.3 & 18.47 & 64.15 & 41.92 & 46.39 & 66.55 & 63.88 \\
  CoCoOp  & 71.02 & \textbf{94.43} & 90.14 & 65.32 & \textbf{71.88} & 86.06 & 22.94 & 67.36 & \textbf{45.73} & 45.37 & 68.21 & 65.74 \\
  \midrule
  \rowcolor{tabhighlight} MuDPT & \textbf{72.13} & 94.07 & \textbf{90.74} & \textbf{65.58} & 71.8 & \textbf{86.13} & \textbf{24.12} & \textbf{67.38} & 45.3 & \textbf{47.14} & \textbf{69} & \textbf{66.13} \\
  MuDPT vs. CoCoOp & \textbf{+1.11} & -0.36 & +0.6 & +0.26 & -0.08 & +0.07 & +1.18 & +0.02 & -0.43 & +1.77 & +0.79 & +0.39 \\ 
  \bottomrule
  \end{tabular}}
\end{table*}
We evaluate our approach under two benchmark settings, \ie, (1) few-shot visual recognition in Sec. \ref{sec:imcls}, and (2) out-of-domain generalization in Sec. \ref{sec:dg}. 

\subsection{Few-shot Visual Recognition}
\label{sec:imcls}
In this section, we measure the visual recognition performance of different prompt tuning methods 
in few-shot setting. We compare our approach \textbf{MuDPT} with zero-shot CLIP \cite{b4}, 
CoOp \cite{b8}, and CoCoOp (current SOTA) \cite{b10}. 

\textbf{Implementation}. we randomly sample a few shot training and validation sets for each dataset while using the original test set for testing where the number of shots is set to 16. We realize our approach based on the best available vision backbone in CLIP, \ie, ViT-B/16. We set the prompt length of 4 for both modalities and initialize the text prompt parameters in the embedding layer using the word embedding of "a photo of a" while random initialization for all other prompts. For visual recognition, all models are trained for 10 epochs with a batch size of 4, prompt tuning depth of 12, and learning rate of 2.5E-3 via SGD optimizer. 

\textbf{Results}. Table \ref{tab:imcls} summarizes the results. In general, Our MuDPT achieves consistent performance improvement than existing text prompt tuning method CoOp \cite{b8}, while obtaining clear advantages in 8 of 11 datasets than CoCoOp \cite{b10}. Specifically, MuDPT obtains 8.2\% accuracy improvements compared with CoOp on average, where the highest accuracy gain is 14.06\% in FGVCAircraft, and the lowest increase is 3.37\% in Flowers102. Averagely, MuDPT achieves 6.31\% accuracy increase over the CoCoOp. To sum up, the experimental results demonstrate the effectiveness of our approach MuDPT, suggests that \textbf{multi-modal deep-symphysis prompt tuning is better than uni-modal prompt tuning}. 

\subsection{Domain Generalization}
\label{sec:dg}
Since prompt tuning on a specific dataset risks learning detrimental disturbance in text and image representations which may impede the generalization ability on out-of-distribution (OOD) data, as suggested in \cite{b10}. In this section, we aim to evaluate the robustness of \textbf{MuDPT} to distribution shift. 
\begin{table}[h!]
  \centering
  \caption{Comparison of MuDPT with existing approaches on base-to-new generalization. 
  Overall, MuDPT achieves consistent improvements compared with CoOp and CoCoOp 
  (current SOTA). }
  \label{tab:base2new}
  \resizebox{0.65\linewidth}{!}{%
  \begin{tabular}{l ccc}
  \toprule
  \textbf{Method} & base & new & \emph{Average} \\
  \midrule
  CLIP      & 69.34 & 74.22 & 71.70 \\
  CoOp      & 82.69 & 63.22 & 71.66 \\
  CoCoOp    & 80.47 & 71.69 & 75.83 \\
  \midrule
  \rowcolor{tabhighlight} MuDPT & \textbf{82.71} & \textbf{74.5} & \textbf{78.39} \\
  MuDPT vs. CoCoOp                              & +2.24 & +2.81 & \textbf{+2.56} \\
  \bottomrule
  \end{tabular}}
\end{table}

\begin{figure*}[ht]
\centering
\includegraphics[width=0.8\linewidth]{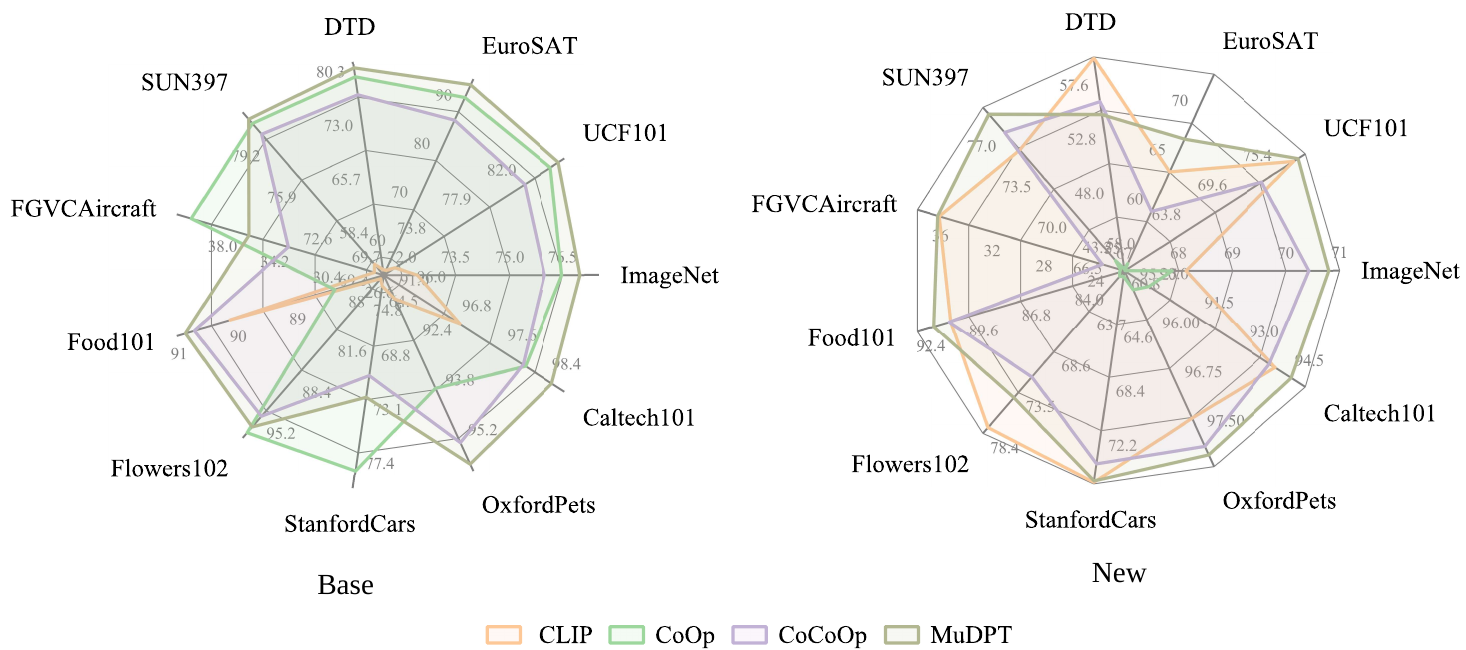}
\caption{Base-to-new generalization results. We compare MuDPT with zero shot CLIP, CoOp, CoCoOp (current SOTA). On base classes, MuDPT outperforms CoCoOp on all 11 datasets. On new classes, MuDPT outperforms CoOp and CoCoOp on all 11 datasets with an exception of 1.13\% accuracy decrease on DTD dataset than CoCoOp.}
\label{fig:base2new}
\end{figure*}

\textbf{Base-to-new generalization}. 
For Base-to-new generalization, we follow CoCoOp \cite{b10} to divide the classes equally into two groups, one as base set and the other as new set. The base set is used for training while testing the performance of the resulting model on the new set. Fig. \ref{fig:base2new} presents the comparison results in base-to-new generalization performance on 11 visual recognition datasets. On base classes, MuDPT shows consistently improved performance on all datasets in comparison with CoCoOp and 8 of 11 datasets better than CoOp. On new classes, we observe that MuDPT outperforms CoOp and CoCoOp on all 11 datasets with an exception of a little accuracy decrease on DTD dataset than CoCoOp. Succinctly, We record the average accuracy on the base and new classes in Table \ref{tab:base2new}. With multi-modal deep-symphysis prompt tuning, MuDPT better fits the base classes and generalizes to new categories compared with CoOp and CoCoOp, where we obtain absolute improvements of 2.56\% in new classes with the best average base accuracy of 82.71\%. Overall, this demonstrates that MuDPT has better generalization ability from base classes to new classes. 

\textbf{Cross-dataset evaluation}. 
We evaluate the cross-dataset generalization ability of MuDPT by training multi-modal deep-symphysis prompts on ImageNet and transferring the learned prompts to the other 10 datasets. Table \ref{tab:xd} shows the comparison of MuDPT, CoOp and CoCoOp. On the source dataset: ImageNet, MuDPT achieves 1.11\% accuracy increase compared to the CoCoOp. Over the target datasets, MuDPT demonstrates better generalization by outperforming CoOp in all datasets and CoCoOp in 7 of 10 datasets. Overall, MuDPT shows better balanced performance between source and target datasets, indicating competitive generalization ability leading to the highest accuracy of 66.13\% averaged over the target datasets.  

\begin{table}[h!]
  \centering
  \caption{Comparison of MuDPT with existing approaches in domain generalization. MuDPT shows highest performance on 3 of 4 target datasets.}
  \label{tab:robustness}
  \resizebox{0.95\linewidth}{!}{
  \begin{tabular}{l ccccc}
  \toprule
  & \textbf{Source} & \multicolumn{4}{c}{\textbf{Target}} \\ \cmidrule(lr){2-2} \cmidrule(lr){3-6}
   & ImageNet &  -V2 & -Sketch & -A & -R \\
  \midrule
  CLIP    &  66.7 & 60.83 & 46.15  & 47.77           & 73.96 \\
  CoOp    &  71.51 & 64.20 & 47.99  & 49.71           & 75.21  \\
  CoCoOp  &  71.02 & 64.07 & 48.75  & \textbf{50.63}  & 76.18  \\
  \midrule
  \rowcolor{tabhighlight} MuDPT & \textbf{72.13} & \textbf{64.92} & \textbf{49.31} & 49.61  & \textbf{77.02} \\
  MuDPT vs. CoCoOp       & \textbf{+1.11} & +0.85 & +0.56 & -1.02 & +0.84 \\
  \bottomrule
  \end{tabular}}
\end{table}
\textbf{Domain generalization}. 
We measure the generalization ability of MuDPT to OOD data by transferring the prompts learned on ImageNet to the other compatible datasets: ImageNet-V2, ImageNet-Sketch, ImageNet-A, and ImageNet-R. We observe that MuDPT obtains better generalization on 3 of 4 datasets as indicated in Table \ref{tab:robustness}. While obtaining the highest accuracy of 72.13 on the source dataset, we achieve accuracy improvements of 0.85/0.56/0.84 on ImageNet-V2, ImageNet-Sketch, ImageNet-R separately. This demonstrates that \textbf{MuDPT is a robust prompt tuning approach among the existing methods in domain generalization setting thanks to multi-modal deep-symphysis prompts fusion}.

\section{conclusion}
In this paper, we identify a critical problem in the existing prompt tuning approaches: the uni-modal design breaks the original alignment state of textual and visual representations in the pre-trained model, which results in sub-optimal performance on downstream tasks.To solve the problem, we propose \textbf{MuDPT} to allow multi-modal deep hierarchical prompt fusion in prompt tuning. In comparison with baselines, Our approach improves the few-shot visual recognition performance and obtains a better trade-off of generalization ability on out-of-domain generalization tasks. However, a limitation of MuDPT is that on 3 of 11 datasets (see Fig. \ref{fig:base2new} (right)), MuDPT's performance still lags behind zero shot CLIP, indicating that more efforts are needed to narrow the gap between continuous prompt tuning and hand-crafted prompts adapted in VL-PTMs in future work. 

\section*{Acknowledgment}
This work was supported by the National Key Research and Development Project of China (No. 2021ZD0110700).


\vspace{12pt}

\end{document}